\documentclass[letterpaper]{article} 
\usepackage[draft]{aaai2026}  
\usepackage{times}  
\usepackage{helvet}  
\usepackage{courier}  
\usepackage[hyphens]{url}  
\usepackage{graphicx} 
\usepackage[numbers]{natbib}  
\usepackage{caption} 
\usepackage{algorithm}
\usepackage{algorithmic}

\usepackage{booktabs}
\usepackage{colortbl}
\usepackage{array}
\usepackage{ragged2e}

\usepackage{amsmath}
\usepackage{xspace}
\usepackage[table,dvipsnames]{xcolor}
\usepackage{amsfonts}
\usepackage{amssymb}
\usepackage{pifont}
\usepackage{longtable}
\usepackage{makecell}

\usepackage{subfig}
\usepackage{overpic} 

\usepackage{multirow}
\usepackage{cellspace}

\usepackage{adjustbox}
\usepackage{subcaption}

\usepackage{newfloat}
\usepackage{listings}
\DeclareCaptionStyle{ruled}{labelfont=normalfont,labelsep=colon,strut=off} 
\floatstyle{ruled}
\newfloat{listing}{tb}{lst}{}
\floatname{listing}{Listing}
\makeatletter

\title{\vspace{-20pt}LSD{\raisebox{-2pt}{\includegraphics[height=13pt]{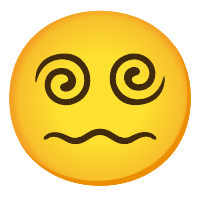}}}-3D: \underline{L}arge-\underline{S}cale \underline{3D} \underline{D}riving Scene Generation with Geometry Grounding}

\author {
    Julian Ost\textsuperscript{\rm 1}\footnote{Indicates equal contribution.},\;
    Andrea Ramazzina\textsuperscript{\rm 2}\footnotemark[1]{},\;
    Amogh Joshi\textsuperscript{\rm 1}\footnotemark[1]{},\;
    Maximilian B\"{o}mer\textsuperscript{\rm 3}, \\
    Mario Bijelic\textsuperscript{\rm 1, 3},\;
    Felix Heide\textsuperscript{\rm 1, 3}
}
\affiliations {
    \textsuperscript{\rm 1}Princeton University
    \quad
    \textsuperscript{\rm 2}Mercedes-Benz
    \quad
    \textsuperscript{\rm 3}Torc Robotics 
}

\usepackage[colorlinks=true,linkcolor=blue,citecolor=blue,urlcolor=blue]{hyperref}

\begin{document}

\maketitle
\begin{abstract}
Large-scale scene data is essential for training and testing in robot learning. Neural reconstruction methods have promised the capability of reconstructing large physically-grounded outdoor scenes from captured sensor data.
However, these methods have baked-in static environments and only allow for limited scene control -- they are functionally constrained in scene and trajectory diversity by the captures from which they are reconstructed.
In contrast, generating driving data with recent image or video diffusion models offers control, however, at the cost of geometry grounding and causality.
In this work, we aim to bridge this gap and present a method that directly generates large-scale 3D driving scenes with accurate geometry, allowing for causal novel view synthesis with object permanence and explicit 3D geometry estimation.
The proposed method combines the generation of a proxy geometry and environment representation with score distillation from learned 2D image priors.
We find that this approach allows for high controllability, enabling the prompt-guided geometry and high-fidelity texture and structure that can be conditioned on map layouts -- producing realistic and geometrically consistent 3D generations of complex driving scenes. Project webpage: \url{https://light.princeton.edu/LSD-3D}.
\end{abstract}    

\section{Introduction}\label{sec:intro}

Large-scale public datasets have driven significant advancements in robot learning over the last decade.
For autonomous driving, large volumes of data~\cite{geiger2012kitti, cordts2016cityscapes, sun2020wod, liao2022kitti360, ettinger2021wod_motion, nuscenes} have unlocked new capabilities for perception and planning.
Initially limited to a few sparsely labeled scenes~\cite{geiger2012kitti}, existing datasets now offer thousands of multi-modal, fully annotated scenes from cities around the world~\cite{nuscenes, sun2020wod} -- unlocking broader capabilities for robot learning. 
However, achieving generalized robot autonomy demands even greater scales of both \textit{diverse} and \textit{complete} data, to capture the long tail of driving scene distributions -- a challenge given the high costs of capturing and annotating real data.
Research in various layers of the robotics stack has demonstrated that training models on large quantities of data and in simulated environments~\cite{dosovitskiy2017carla,waymax2023,borkman2021unity,rong2020lgsvl, shah2018airsim,talwar2020evaluating} can result in robust and more generalized autonomy.

Recent work in neural scene reconstruction promises to bridge this gap by reconstructing previously simulated driving environments directly from sensor data~\cite{ost2021neural,tonderski2023neurad,yang2023unisim_neuralclosedloop,wu2023mars}.
Such scenes can be replayed in real time, synthesize novel views, and allow for unseen actor variations in closed-loop testing~\cite{tonderski2023neurad,ljungbergh2024neuroncap,Dauner2024NEURIPS}. However, neural reconstruction methods are fundamentally limited in that they cannot produce novel content beyond recorded scenes -- they do not offer data-driven simulation with great scene diversity.

Video diffusion models have been proposed to increase data volumes and diversity. Pretrained on internet-scale datasets and subsequently fine-tuned on autonomous driving data they generate videos which mimic driving datasets and multi-view camera setups~\cite{wang2023drivedreamer,lu2023wovogen,gao2024vista, nvidia2025cosmosdrivedreams}. While these methods can generate a large corpus of novel driving data, outside of their training data, and provide feature control, they also come with inherent limitations. High computational costs, on the order of seconds per generated multi-view frame, restrict real-time replay and scalability for closed-loop simulation tasks. Furthermore, they lack explicit spatial modeling, which prevents causality, object permanence, and 3D consistency. The latter also prohibits them from replaying novel trajectories within a pregenerated environment~\cite{gao2024magicdrive3d}. As such, video diffusion models struggle as data-driven simulators, especially for safe robot learning. 

Explicit 3D scene models \textit{guarantee causality} and 3D consistency. Directly generating explicit 3D scenes, however, poses a challenge: both \textit{geometry} and \textit{texture} have to be generated consistently and with high quality.
Some approaches use LiDAR point clouds to produce pure geometry without detailed texture~\cite{lee2024semcity, zhang2024urbandiff,ren2024xcube, zyrianov2024lidardm} -- which cannot be used to train image-based autonomous driving models -- while more recent approaches adopt coarse 3D geometry as a conditioning mechanism for video synthesis ~\cite{lu2023wovogen, lu2024infinicube}. As such, these methods inherit video diffusion limitations, such as a lack of causality. 
Distillation methods~\cite{poole2022dreamfusion} instead address the 3D data gap by transferring priors from 2D image models into 3D representations via inverse rendering techniques~\cite{mildenhall2021nerf,kerbl3Dgaussians, yi2023gaussiandreamer}.
However, existing techniques are limited to object-centric generation~\cite{poole2022dreamfusion, xie2024latte3d, trellis} and lack realism~\cite{li2024dreamscene} –– and so, they are not suitable for the complexity~\cite{zhang20243d} or spatial scale~\cite{shriram2024realmdreamer} of large-scale driving scenes. 
Only recent work has explored distillation approaches for scene extrapolation from sparse real data captures~\cite{wu2025difix3d, wu2024reconfusion,liu20243dgsEnhancer,fischer2025flowr}, achieving improved reconstruction quality from few images and hinting at the potential for complete outdoor scene generation.

We propose a novel approach that overcomes the aforementioned limitations of existing generative methods for large-scale driving scenes. Our method generates explicit 3D models of entirely novel environments -- with both geometry and high-fidelity texture -- by fusing the diversity of image and video generation with the efficiency of explicit 3D representations.
We first generate a coarse geometry of street scenes, optionally controlled by a road map layout.
This proxy is then used to guide the generation of fine structural details and high-fidelity textures via image space distillation with a high-quality image generation model.
Specifically, we introduce geometry-grounded distillation sampling (GGDS), an image space sampling approach, that incorporates explicit geometry control and exact noise sampling by DDIM inversion in a single method. We find that the combination of geometry guidance and consistent noise sampling through inversion can deliver successful and 3D consistent scene generation via distillation.
The method produces diverse, realistic, and large-scale 3D scene models for autonomous driving.
Generated scenes guarantee causality and can produce unlimited novel trajectories in real-time -- enabling \textit{scalability} -- while maintaining 3D consistency and appearance fidelity.
Furthermore, precise prompt control over weather, season, time-of-day, and location, in the form of explicit environmental lighting, enables further fine-grained customization of these virtual scenes.

We validate our method on the Waymo Open Dataset~\cite{sun2020wod}, generating novel scenes which not only inherit the data prior distribution, but leverage the implicit prior of 2D diffusion models to provide enhanced scene \textit{diversity}.
Our approach, generating complete and coherent large-scale 3D scenes, outperforms state-of-the-art existing generative methods in image synthesis of unseen camera angles by \textbf{18\% in FVD} and maintains prompt adherence on the level of pure video-based approaches.

We summarize our specific contributions as follows:
\begin{itemize}
    \item To our knowledge, our method is the first to utilize a distillation approach to directly \emph{generate} and optimize explicit 3D driving scenes with high-quality geometry and texture -- guaranteeing causal generation.
    \item We introduce Geometry-Grounded Distillation Sampling (GGDS), a method combining controlled proxy mesh generation with a conditional diffusion prior to produce novel, view-consistent Gaussian splatting scenes, with real-time rendering and composability with 3D assets.
    \item We generate diverse large-scale scenes which can be rendered into physically-grounded videos controlled by trajectories through the scene, enabling the creation of unlimited, completely unseen environments, controlled by scene descriptions, traffic map layouts, or text prompts.
\end{itemize}


\newcommand{\redcheck}{{\color{red}\ding{55}}\xspace}
\newcommand{\red}{\cellcolor{red!12.5}\footnotesize{\redcheck}}
\newcommand{\yellowcheck}{{\color{YellowOrange}(\ding{51})}\xspace}
\newcommand{\yellow}{\cellcolor{YellowOrange!12.5}\footnotesize{\yellowcheck}}
\newcommand{\greencheck}{{\color{Green}\ding{51}}\xspace}
\newcommand{\green}{\cellcolor{Green!12.5}\footnotesize{\greencheck}}

\begin{table}[t!]
\centering
\resizebox{\linewidth}{!}{
\begin{tabular}{p{0.24\textwidth} | c c c c c c c c}

\Xcline{1-9}{3\arrayrulewidth}

\footnotesize{Method} & \footnotesize{Unlimited}  & \footnotesize{Compos-} & \footnotesize{Causal 3D} & \footnotesize{Real-Time} & \footnotesize{View} &  \multicolumn{3}{c}{\footnotesize{Control}} \\
\Xcline{7-9}{3\arrayrulewidth}
\footnotesize{} & \footnotesize{Viewpoints} & \footnotesize{ability} & \footnotesize{Geometry} & \footnotesize{Rendering} & \footnotesize{Extrapolation}  & \footnotesize{Weather} & \footnotesize{Time} & \footnotesize{Map} \\
\hline
\footnotesize{DriveDreamer}~\cite{wang2023drivedreamer} & \red & \red & \red & \red & \red & \green & \green &  \red \\

\footnotesize{WonderJourney}~\cite{yu2024wonderjourney}  & \red & \red & \red &  \red &  \red & \green & \green & \red   \\ 

\footnotesize{Streetscapes}~\cite{deng2024streetscapes} & \red & \red & \red & \red & \red & \green & \green & \green \\

\footnotesize{Vista}~\cite{gao2024vista} & \red & \red & \red & \red & \red & \green & \green &  \red \\

\footnotesize{MagicDriveDiT}~\cite{gao2024magicdrivedit} & \red & \red & \red & \red & \red & \green & \green &  \red \\

\footnotesize{WoVoGen}~\cite{lu2023wovogen} & \red & \red & \red & \red & \red & \green & \green & \green \\

\footnotesize{WonderWorld}~\cite{yu2024wonderworld}  & \red & \red & \red &  \red &  \red & \green & \green & \red   \\ 

\hline

\footnotesize{NF-LDM}~\cite{kim2023nfldm} & \red & \red & \red & \green & \red  & \green & \green & \green \\

\footnotesize{InfiniCity}~\cite{lin2023infinicity} & \red & \red & \green & \green & \green & \red & \red & \green \\

\footnotesize{MagicDrive3D}~\cite{gao2024vista} & \red & \green & \yellow  & \green & \green & \green & \green & \green \\

\footnotesize{CityDreamer}~\cite{xie2024citydreamer} & \red & \red & \green & \green & \green & \red & \red & \green \\

\footnotesize{InfiniCube}~\cite{lu2024infinicube} & \red & \green & \yellow & \green  & \green & \green & \green & \green \\

\footnotesize{GEN3C}~\cite{ren2025gen3c}  & \red & \red & \yellow &  \red &  \green & \green & \green & \red   \\

\footnotesize{\textbf{LSD{\raisebox{-2pt}{\includegraphics[height=10pt]{fig/emoji.pdf}}}-3D} (\textbf{Ours})} & \green & \green & \green &  \green & \green &  \green & \green & \green \\
\Xcline{1-9}{3\arrayrulewidth}
\end{tabular}
}

\caption{\textbf{Video and 3D Generation for Driving Scenes}. We review the recent body of video (top) and 3D generation work (bottom). \textit{Please zoom in digital document for details.}}
\label{tab:capabilities}
\vspace{-12pt}
\end{table}




\section{Related Work}\label{sec:related}
\paragraph{Image and Video Synthesis.}
Recent advances in image generation have enabled the synthesis of high-resolution, photorealistic imagery. 
These approaches -- from generative adversarial networks (GANs)~\cite{goodfellow2014generative, brock2018large, karras2020analyzing, sauer2022stylegan}, likelihood-based methods~\cite{kingma2013auto, razavi2019generating, vahdat2020nvae}, to more prominently, diffusion models~\cite{dhariwal2021diffusion, song2020denoising, dockhorn2021score, esser2021taming, ho2020denoising, rombach2022high, vahdat2021score} -- have been recently extended to video generation~\cite{blattmann2023stablevideo, zhang2024trip, singer2022make, rombach2022high, harvey2022flexible, ho2022imagen, ho2022video, gao2024magicdrivedit, yang2024cogvideox, hong2022cogvideo, genmo2024mochi}.
Early methods offer control via text prompt~\cite{singer2022make}, image~\cite{zhang2024trip}, or camera position~\cite{sargent2023zeronvs, yang2024direct}.
Recent methods like GEN3C~\cite{ren2025gen3c, team2025aether} or GeoDrive~\cite{chen2025geodrive3dgeometryinformeddriving} for driving scenes incorporate 3D conditioning and point projection to improve geometric consistency.
Specialized models, such as Vista~\cite{gao2024vista}, MagicDrive~\cite{gao2023magicdrive, gao2024magicdrivedit}, or DriveDreamer~\cite{wang2023drivedreamer}, have been developed on top of foundational video models~\cite{blattmann2023stablevideo} for autonomous driving applications -- employing bounding box, HD-map, and dense voxel guidance for feature control.
These methods extend pre-trained video diffusion models to generate videos which mimic driving sensor setups.
However, despite efficiency improvements~\cite{song2024sdxs, luo2023latent}, video generation models remain computationally intensive~\cite{yang2024cogvideox, genmo2024mochi, wan2025wan} and preclude the scalability necessary for real-time simulation. 
Furthermore, they struggle with coherent novel-view rendering over long driving trajectories and suffer from a lack of causality.
In the top section of Tab.~\ref{tab:capabilities}, we provide a comparison of relevant video synthesis methods for large-scale driving scenes.

\paragraph{3D Generation.}
To guarantee consistency and causality, recent approaches separate from 2D image and video generation and focus on the explicit generation of individual, object-centric 3D structures.
These methods~\cite{trellis,li2025triposg,wu2024cat4d,gao2024cat3d,jun2023shape,sargent2023zeronvs,hunyuan3d22025tencent,xie2024latte3d} generate consistent 3D structure by leveraging well-defined features and latent spaces, and multi-view observations -- enabling them to directly perform explicit diffusion of 3D objects.
However, to train these 3D generative models, they rely on high-quality synthetic 3D data~\cite{objaverseXL, collins2022abo}, and most works are thus constrained to the generation of individual objects -- generation of large-scale 3D outdoor scenes remains an open challenge.

\begin{figure*}[thb!]
    \centering
    \includegraphics[width=0.95\linewidth]{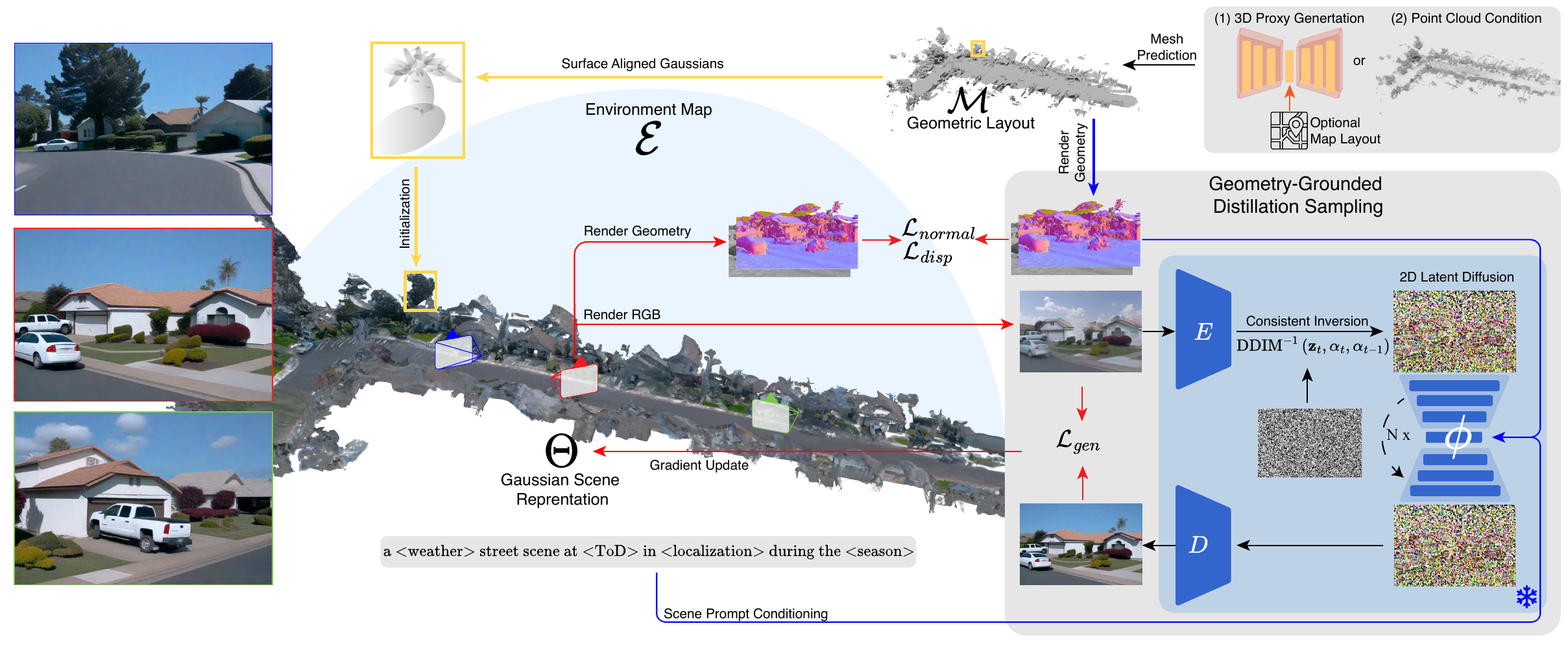}
    \caption{\textbf{Geometry-Grounded Large-Scale 3D Scene Generation.}
    We generate a large-scale scene as a combination of a coarse geometric layout, an environment map, and a set of Gaussians for texture details, discussed in Sec.~\ref{ssec:representation}.
    The geometric layout is either generated, conditioned on a map, or predicted from point-cloud data and guides the overall scene structure.
    We can further control the setting with a scene prompt, describing time-of-day, season, and weather.
    Through Geometry-Grounded Distillation Sampling (GGDS), we then further optimize the Gaussian-based scene representation by leveraging 2D priors from the conditional latent diffusion model through consistent diffusion sampling and image-space optimization -- together with a set of geometry-grounding regularizers -- and generate a causal large-scale scene representation as described in Sec.~\ref{ssec:optimization}.}
    \label{fig:overview}
\end{figure*}

\paragraph{Scene Reconstruction and Generation.}
For large-scale scene generation~\cite{kim2023nfldm,yang2023urbangiraffe,lee2024semcity,anciukevicius2022renderdiffusion,wang2023taming,bahmani20234d, lin2023infinicity}, various 3D representations have been proposed: triplanes~\cite{shue20233d,lee2024semcity,anciukevicius2022renderdiffusion}, semantic occupancy grids, bounding boxes, and 3D maps~\cite{yang2023urbangiraffe,xu2023discoscene}. However, reliance on generation of explicit priors requires expensive annotation data, and these early methods' generations lack significantly in photorealism and scale. 
In contrast, satellite imagery has been used for the city-scale 3D generation of urban environments~\cite{lin2023infinicity,xie2024citydreamer}.
More recently, hierarchical voxel diffusion methods~\cite{ren2024xcube, ren2024scube}, have used accumulated LiDAR point clouds to supervise the generation of 3D driving scene meshes or directly generate realistic LiDAR point clouds~\cite{zyrianov2024lidardm}. 
However, these methods exclusively generate geometry, without the texture or appearance needed for training perception models in simulation -- see the bottom section of Tab.~\ref{tab:capabilities}.

Neural scene reconstruction~\cite{mildenhall2021nerf} on the other side has shown promising results for the production of high-quality, photorealistic visual data. 3D Gaussians~\cite{kerbl3Dgaussians} have emerged as prime scene representations, able to explicitly model geometry while also allowing for real-time rendering, enabling scalability.
Methods such as OmniRe~\cite{chen2025omnire}, SplatAD~\cite{hess2025splatad}, SCube~\cite{ren2024scube} or STORM~\cite{yang2025storm} are capable of reconstructing 3D texture and geometry as 3D Gaussians from real-world driving videos, allowing for the exploration of novel trajectories.
Nevertheless, pure reconstruction methods are still fundamentally limited by the availability of real data to reconstruct from.
A natural extension of these works has been undertaken by works like WoVoGen~\cite{lu2023wovogen} and InfiniCube~\cite{lu2024infinicube}, which replace the real data required for scene reconstruction with generated videos conditioned on scene geometry -- the latter then fits these videos onto a set of deformable Gaussian Splats, a first approach in 3D grounding.
However, this approach is still inherently limited to the generated video trajectory.
Furthermore, despite the explicit 3D representation, the lack of causality in a single original video results in visual artifacts and inconsistencies being baked into the scene representation.
In contrast, distillation is a paradigm which has recently emerged in 3D object generation~\cite{poole2022dreamfusion, anciukevicius2022renderdiffusion,liang2024luciddreamer,miao2024dreamerxl} and sparse reconstruction~\cite{wu2025difix3d, wu2024reconfusion,liu20243dgsEnhancer,fischer2025flowr,gao2024cat3d} that is centered around the learning of a neural 3D representation guided by pretrained text-to-image models. 
This approach makes it possible to synthesize novel 2D views from arbitrary camera positions, while ensuring 3D consistency and visual fidelity throughout the optimization, thereby bypassing the problem of limited 3D data availability. However, generation by distillation works for object-centric views but does not naturally scale to complex textures and large-scale scenes, as is necessary for driving scene generation.
Our work investigates how distillation can be extended for large-scale driving scene generation, allowing for the \textit{distillation} of image priors as part of the 3D generation -- as opposed to explicit \textit{supervision}.

\newcommand{\norm}[1]{\left\lVert#1\right\rVert}
\begin{figure*}[thb!]
    \centering
    \includegraphics[width=0.98\linewidth]{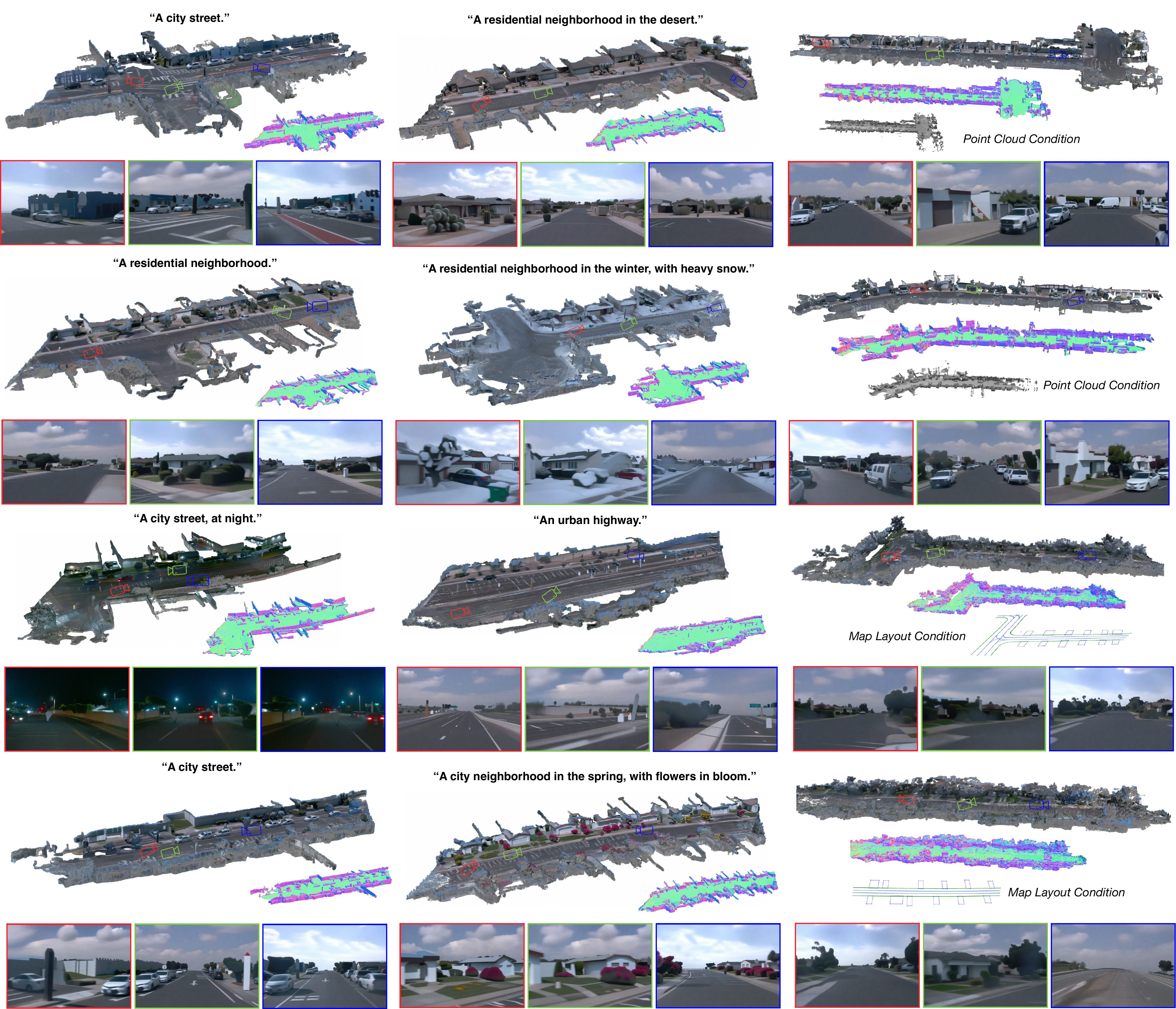}
    \caption{\textbf{Geometry-Grounded LSD{\raisebox{-2pt}{\includegraphics[height=11pt]{fig/emoji.pdf}}}-3D Generations}. We visualize 3D scenes generated via our method, alongside the corresponding map of surface normals and selection of novel viewpoints at street level for each of them. In the first two columns, we provide samples of scenes with diversity in time-of-day, season, location, and scene type. In the third column, we provide examples of generated scenes with a point cloud condition and map layout condition. We confirm the method generates diverse, explicit, and causal 3D scenes.}
\label{fig:results}
\end{figure*}

\section{Geometry-Grounded 3D Generation}\label{sec:method}
Our method generates large-scale driving scenes with 3D-consistent geometry and texture.
We provide an overview of the generative process in Fig.~\ref{fig:overview}. In the following, we first describe our large-scale scene representation before introducing the proposed conditional generation process.

\subsection{Scene Representation}\label{ssec:representation}
\paragraph{Geometric Layout and Background Environment.}
The coarse geometric layout of the scene encodes road, rough vegetation, static vehicles, and building facades. The layout is represented by a mesh $\mathcal{M} = \{ \mathbf{F_1},..., \mathbf{F_N} \}$ where each triangular face is defined by three vertices $\mathbf{F} = [ \textbf{V}_a , \textbf{V}_b, \textbf{V}_c ], \textbf{V} \in \mathbb{R}^3$.

We model the background texture at infinity with an environment map~\cite{greene1986environment}.
For a given time of day, weather, and seasonal setting, we introduce into our scenes a corresponding background environment in the form of an equirectangular map $\mathcal{E}$ of size $ H_{env} \times W_{env} \times 3$, that offers explicit environment lighting control.
Queried as a spherical environment map $f_{env} \left(\textbf{d}, \mathcal{E}\right)$, it returns a color $\mathbf{c}$ for any given viewing directions $\textbf{d} = (\varphi, \eta) \in \mathbb{R}, (0, 2\pi ]$.

\paragraph{Gaussian Structure and Texture.}
On a finer level, we represent detailed foreground geometry and texture as a set of 2D oriented planar Splats introduced by Huang \emph{et al.}~\shortcite{Huang20242DGS}.
Each splat $\theta_k$ is parametrized by its central point $\mathbf{p}_k \in \mathbb{R}^3$, two principal tangential vectors $\mathbf{t} = (\mathbf{t_u}, \mathbf{t_v})$ that define their orientation, and a variance controlling its scale per axis $\mathbf{s}=(s_u , s_v)$.
Our complete scene representation $\Theta$ is defined as the set of all $K$ individual Gaussian $\Theta = \left[\theta_0, ...,\theta_K \right]$.
Complex textures are further modeled by the Gaussian appearance $\mathbf{c}$ (stored as a set of spherical harmonics) and opacity $\mathbf{o}$.
The rendered foreground is alpha-composited with the environment map rendered at infinity for each pixel's viewing direction.

\subsection{Geometric Layout Generation}\label{ssec:coarse}

\begin{figure*}
    \centering
    \includegraphics[width=\linewidth]{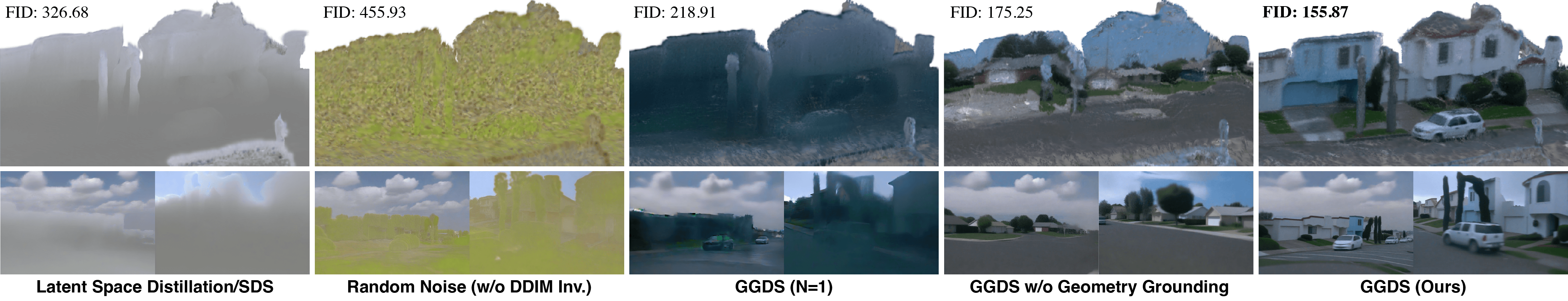}
    \caption{\textbf{Ablation Experiments}. We qualitatively validate the core components of our optimization method. With vanilla SDS, scenes completely fail to converge, necessitating our {Gaussian Optimization} approach. The proposed texture regularization and initialization approach ensures that scenes converge to a reasonable color distribution, while scenes without them fail. The bottom table reports FID scores without the same components, quantitatively confirming that optimization, geometry-grounding, and texture regularization prove critical for producing high-quality 3D driving scenes.
    }
    \label{fig:ablation}
\end{figure*}

To generate the foreground mesh geometry, we first generate a voxel occupancy $V$ from a hierarchical latent voxel diffusion model~\cite{rombach2022high}, which makes use of a dense, low-resolution and a sparse, high-resolution 3D UNet~\cite{Zhou_2021_ICCV,ren2024xcube} as its respective backbones (see the Supplementary Material for details).
To enable explicit control of the layout of $V$, we condition the diffusion model on a map layout $M$ to model the conditional probability $p \left(V | M \right)$.
As single-scene generations are limited to $100m \times 100m$, we expand the generation size by introducing chunk-wise outpainting.
The initial chunk is exclusively conditioned on the locally aligned map $M_{e}$, but each subsequent chunk $e$ is also conditioned on an overlapping zone with the previous chunk $e-1$ with $p \left(V_e | M_{e}, V_{e-1} \right)$.
This diffusion model is trained from scratch on the aggregated point-cloud data and maps of the target street scene dataset~\cite{sun2020wod}.
From the generated voxel grid, we then predict the enclosing coarse surface mesh geometry $\mathcal{M}$ with neural kernel surface reconstruction~\cite{huang2023nksr,williams2021nkf}.

\subsection{Geometry-Grounded Scene Generation}\label{ssec:optimization}
Given the generated coarse mesh $\mathcal{M}$ and a monochromatically initialized environment map $\mathcal{E}$, we next generate a textured scene with causal consistency.

\paragraph{Mesh to Gaussian Representation.}
We place Gaussians $\Theta$ at $\textbf{p}_k$, to represent mesh faces $\mathbf{F}$ and set orientation, scale, and tangential axes according to the triangle normal $\textbf{n}_F$, and area $a_F$, with orientation $\textbf{n}_F = \textbf{t}_u \times \textbf{t}_v$ and scale $| \mathbf{s} | = a_F$.

\paragraph{Geometry-Grounded Distillation Sampling (GGDS).}

We next distill a latent diffusion model (LDM) $p_{\phi , data} \left( \mathbf{z} \right)$ on the set of Gaussians $\Theta$ through a novel iterative optimization method, which we term Geometry-Grounded Distillation Sampling (GGDS).
This optimization method is designed to avoid the artifacts that are typically present in existing latent space distillation-based models~\cite{poole2022dreamfusion, liang2024luciddreamer, hwang2024vegs} and in ego-centric scenes (see Fig.~\ref{fig:ablation}).

In each distillation step, we first obtain an image $x_i = g \left(\Theta, \psi_i \right)$ for viewpoint $\psi_i$ with the rasterization function $g$.
We encode the latent $z_{0,i} = E\left( x_i \right)$ from this image and add noise $\epsilon$ of noise level $t$ to obtain the noisy latent $z_{t}$.
The noise level $t$ is sampled uniformly between $t_{max}$ and $t_{min}$.
The noisy latent $z_{t}$ is the denoised for $N$ steps and decoded, generating a ground-truth image $\hat{\mathbf{x}}_i = D\left( \hat{\mathbf{z}}_{0,i} \right)$ for the respective viewpoint, inducing a loss in image-space.
We formulate the objective as the image reconstruction loss between the generated image $\hat{\mathbf{x}}_i$ and the rendered image $\mathbf{x}_i = g \left(\Theta, \psi_i \right)$:
\begin{align}
	\begin{split}
		\mathcal{L}_{gen} \left( \Theta \right) = 
		& \mathbb{E}_{\psi_i, t} \left[
		\omega\left( t \right) \left( 
		\|g \left(\Theta, \psi_i \right) - \hat{\mathbf{x}} \| \right. \right. \\
		& \left. \left. + \mathcal{L}_{LPIPS}(g \left(\Theta, \psi_i \right), \hat{\mathbf{x}})
		\right) \right]
	\end{split}
\end{align}

where $\omega\left( t \right)$ is the noise-level dependent weight and $\mathcal{L}_{LPIPS}$ is the perceptual similarity~\cite{zhang2018perceptual}.
We choose $N = 5$ independently of $t$, which we show allows for higher generation quality for lower noise level $t$ in the later stages of the scene optimization.
To enforce progressive optimization from coarse to fine, the respective generation strength from the image prior is linearly annealed by dropping the lower sampling bound $t_{min}$.
Directly optimizing in image space has significant advantages over score distillation, as our ablation experiments validate, where latent optimization is unable to converge on non-overlapping viewpoints.

\begin{figure*}
    \centering
    \includegraphics[width=\linewidth]{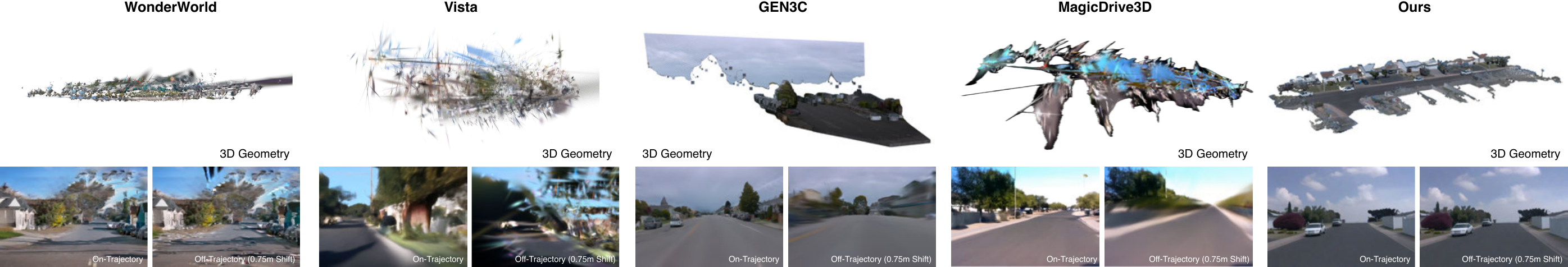}
    \caption{
    \textbf{Qualitative Comparisons to Video and Scene Generation Methods.}
    Our approach generates an accurate and 3D-consistent scene representation, enabling high-quality novel view synthesis and the generation of unlimited off-trajectory viewpoints.
    In contrast, existing baselines WonderWorld~\cite{yu2024wonderworld}, Vista~\cite{gao2024vista} combined with Gaussian Splatting~\cite{Huang20242DGS,kerbl3Dgaussians}, GEN3C~\cite{ren2025gen3c} and MagicDrive3D~\cite{gao2024magicdrive3d}, which generate driving videos and thus lack implicit spatial modeling, fail to generate a consistent and 3D-plausible scene, precluding the production of novel driving trajectories (please zoom into PDF version for details).
    } 
    \label{fig:qualitative}
\end{figure*}

\begin{table*}[t!]
\centering
\renewcommand{\arraystretch}{1.4} 
\resizebox{1.0\linewidth}{!}{
\begin{tabular}{p{0.35\textwidth} | p{0.23\textwidth} | c c c | c c c | c}
\Xcline{1-9}{3\arrayrulewidth}
\multicolumn{1}{c|}{Method} & \multicolumn{1}{c|}{Training / Finetuning} & \multicolumn{3}{c|}{Seen} & \multicolumn{3}{c|}{Novel} & \multicolumn{1}{c}{CLIP}\\
\cline{3-5} \cline{6-8}
& & FID $\downarrow$ &  $\textrm{FD}_{\textrm{DINOv2}}$ $\downarrow$ & FVD $\downarrow$ & FID $\downarrow$ & $\textrm{FD}_{\textrm{DINOv2}}$ $\downarrow$ & FVD $\downarrow$ & {\shortcite{Radford2021LearningTV} $\uparrow$}  \\
\hline
WonderWorld~\shortcite{yu2024wonderworld} + 2DGS~\shortcite{Huang20242DGS} &  WOD~\shortcite{sun2020wod} & 130.17 & 1333.88 & 1315.53 & \underline{220.61} & \underline{1489.51} & 1424.56 & 28.88 \\ 
Vista~\shortcite{gao2024vista} + 2DGS & OpenDV~\shortcite{yang2024genad}/WOD & \underline{111.93} & 1510.25 & 1023.4 & 242.03 & 1805.96 & \underline{1190.76} & 26.51 \\
MagicDrive3D~\shortcite{gao2024vista} \footnotemark[1] & NuScenes~\shortcite{nuscenes}/WOD\footnotemark[2] & 139.11/178.71\footnotemark[2] & 1950.67/1965.31\footnotemark[2] & 2285.5/1585.48\footnotemark[2] & 163.73/186.36\footnotemark[2] & 2004.21 & 1665.30 & 21.03 \\
GEN3C~\shortcite{ren2025gen3c} with Cosmos~\shortcite{nvidia2025cosmosworldfoundationmodel} &  Proprietary/WOD & \textbf{97.71} & \underline{1331.16} & \textbf{962.02} & 273.27 & 2237.03 & 1617.62 & 26.57 \\ 
\textbf{LSD{\raisebox{-2pt}{\includegraphics[height=11pt]{fig/emoji.pdf}}}-3D} Layout + 2DGS Dist. (\textbf{Ours}) & WOD & 119.38 & \textbf{1247.19} & \underline{989.94} & \textbf{119.18} & \textbf{1227.62} & \textbf{974.36} & 26.03 \\
\Xcline{1-9}{3\arrayrulewidth}
\end{tabular}
}
\caption{\textbf{Quantitative Evaluation of Generation Quality for 3D Scene Generation} with our proposed method and existing approaches. Best results in each visual quality category are in \textbf{bold} and the second best \underline{underlined}. We report CLIP scores on the right, confirming that our generations adhere to prompts on par with existing methods. We apply the same pre-trained T2I model~\cite{podell2023sdxl} for all models, and the same 2DGS~\cite{Huang20242DGS} pipeline for optimization, to provide fair comparisons. Significantly lower FVD and FID score on novel views confirm temporal and geometric consistency is achieved by our method through geometrically grounded generations.}
\label{tab:quantitative_results}
\end{table*}








To further mitigate randomness that leads to diverging optimization objectives, we enforce consistency between optimization steps at the same viewpoint through DDIM inversion instead of random noise sampling from noise level ${t}$.
This ensures a higher level of consistency between the rendered $\mathbf{x}_i$ and generated images $\hat{\mathbf{x}}_i$ even in later steps of the optimization, which is in contrast to random sampling, where high noise levels $t$ can lead to extreme disagreement.
We propose a fixed $N$-step DDIM inversion~\cite{song2020denoising} at any noise-level t and directly predict

\begin{equation}\label{eq:fwdDDIM_main}
\begin{aligned}
    \mathbf{z}_{t,i} =& \mathrm{DDIM}^{-1}\left( \mathbf{z}_{t-1, i}, \alpha_{t}, \alpha_{t - 1}\right) \\ 
    =& \frac{\sqrt{\alpha_t}}{\sqrt{\alpha_t-1}}\left( \mathbf{z}_{t-1} - \sqrt{1- \alpha_{t-1}} \epsilon_\Phi (\mathbf{z}_{t-1})\right) \\ &+ \sqrt{1-\alpha_t} \epsilon_\Phi (\mathbf{z}_{t-1}) \textrm{.}
\end{aligned}
\end{equation}

where $\mathbf{z}_t$ and $\mathbf{z}_{t-1}$ represent the noisy latent, $\epsilon_\Phi$ the predicted noise,  and $\{\alpha_t\}_{t=0}^T$ indicate noise level indexing a monotonically increasing time schedule. This allows the model to only introduce changes exactly where needed in each optimization step to satisfy the 2D diffusion prior, also confirmed in Fig.~\ref{fig:ablation}. Given this loss objective, we then optimize the Gaussian scene representation through Stochastic Gradient Langevin Dynamics (SGLD) updates~\cite{kheradmand20243dgaussiansplattingmarkov} with
\begin{equation}\label{update}
    \Theta_{k+1} = \Theta_{k} + \xi \left( \nabla_{\Theta} \mathcal{L}_{\mathrm{gen}} \left( \Theta_{k} \right) \right) + \lambda_{noise} \epsilon \text{,}
\end{equation}
where $\lambda_{noise} \epsilon$ is the perturbation of each Gaussian in $\Theta$.

As text conditioning $c$ alone is not a suitable prior to achieve the style of street scenes, we first finetune a LDM $p_{\phi , fine-tune} \left( \mathbf{x} \right)$ on the desired image distribution. Additionally, to avoid scene geometry drifting from the initial mesh, we incorporate disparity conditioning ~\cite{zhang2023adding} on disparity maps computed from the rendered mesh depth $\mathcal{D}= g_{d}(\mathcal{M}, \psi_i)$ to the denoising process $p_{\phi , fine-tune}\left( x  | c_{text}, g_{d}(\mathcal{M}, \psi_i)\right)$. This step is crucial for consistency across views and the gradient signal from the 2D diffusion prior on the generated proxy geometry.

\paragraph{3D Geometry Loss.}
Alongside geometry conditioning in the 2D diffusion process, we also regularize all $\theta_k$ to retain high-quality, smooth 3D geometry.
This is achieved by penalizing splat orientation of normal maps $\mathcal{N}_{\Theta, \psi_i}$ and disparity $\mathcal{D}_{\Theta, \psi_i}$ from viewpoint $\psi_i$ with respect to $\mathcal{N}_{\mathcal{M}, \psi_i}$ and $\mathcal{D}_{\mathcal{M}, \psi_i}$ rendered from the normals of the proxy geometry as
\begin{equation}
    \mathcal{L}_{norm} = \|\mathcal{N}_{\Theta, \psi_i} - \mathcal{N}_{\mathcal{M}, \psi_i} \| \textrm{ and }
    \mathcal{L}_{disp} = \|\mathcal{D}_{\Theta, \psi_i} - \mathcal{D}_{\mathcal{M}, \psi_i} \|.
\end{equation}
Moreover, we apply Gaussian regularization from Huang et. al~\shortcite{Huang20242DGS}.
Finally, we employ a total-variation (TV) loss on the rendered images to encourage noise reduction in our scenes.
We refer to the Supplemental Material for detailed parameter settings and loss composition.

\paragraph{Deferred Rendering.}
We propose a deferred rendering process for novel trajectory videos, inspired by Thies et al.~\shortcite{thies2019deferred}. After scene generation, we use the Gaussian rasterizer to produce an initial frame $\mathbf{x}_0$ and encode it into a slightly noisy latent $\mathbf{x}_t$, which is deferred into final output $\mathbf{x}_0^{(1)}$ using a fine-tuned T2I model.

This rendering procedure allows us to generate photorealistic images with both low- and high-frequency textures in image space, for instance the road, or a tree, respectively.

\section{Assessment}\label{sec:results}

In this section, we validate our approach both quantitatively and qualitatively. We conduct an ablation study to validate our design choices and investigate the realism of rendered scenes alongside competing baseline methods.

\subsection{Ablation Study}
We validate the effects of key components of our scene optimization method and visualize the results in Fig~\ref{fig:ablation}.
Specifically, we analyze the effect of incorporating our improved geometry-grounded distillation sampling (GGDS) approach with SDS, image-space sampling without DDIM inversion, the effectiveness of multi-step denoising and geometry-conditioned diffusion guidance. 
Without the proposed GGDS, especially in the case of SDS and random noise sampling results, the method experiences catastrophic failure.
This highlights the need for two major design choices, which both result in more consistent optimization and distillation of the underlying 3D scene representation.
Further multi-step diffusion results in reduced scene generation quality of dark scenes.
Without geometry-grounding diffusion guidance, objects in the scene (such as houses or cars) do not follow the conditioning proxy geometry, resulting in inaccurate and flat 3D geometry and poor novel trajectory results. We confirm the visual trend quantitatively in Fig.~\ref{fig:ablation} bottom, which validates that our method (with all proposed components) produces the best visual quality.

\subsection{Experiments}
We validate our method by comparing it against four distinct approaches for 3D scene and video generation. 
\paragraph{Baselines.} We ran two image-to-video generation methods: (a) Vista~\cite{gao2024vista}, a full driving video model trained on internet-scale driving videos, and (b) WonderWorld~\cite{yu2024wonderworld}, in combination with our fine-tuned image model~\cite{podell2023sdxl} on the target dataset~\cite{sun2020wod} for fair comparison.
For both methods, we fit 2D Gaussian Splats~\cite{Huang20242DGS} to assess geometric causality and novel view synthesis quality.

We also test two conceptually different 3D generation baselines: (c) MagicDrive3D~\cite{gao2024magicdrive3d}, which proposes a driving scene-specific multi-view video diffusion and Gaussian reconstruction pipeline, and (d) GEN3C~\cite{ren2025gen3c}, which fuses geometry prediction from a single image with the Cosmos-Predict~\cite{nvidia2025cosmosworldfoundationmodel} video diffusion model.
Due to the unavailability of public code and models of any candidate~\cite{gao2024magicdrive3d,lu2024infinicube} at the time of submission, we rely on our own implementation (Layout + Geometry Controlled Video Generation) with MagicDrive3D built on top of the latest diffusion transformer version of MagicDriveDiT~\cite{gao2024magicdrivedit}.
See the Supplementary Material for details.

\paragraph{Computational Requirements.} Each scene is generated with 6000 steps, corresponding to an average time of 2 hours on a single NVIDIA H100 GPU. Comparable methods~\cite{liang2024luciddreamer} that generate single objects, i.e., dramatically smaller scenes, require similar runtime at 5000 steps.
Our method outputs frames at 960p resolution at rates higher than 60 fps, providing real-time rendering capabilities.
Scenes are initialized with 1.8 to 2.2 million Gaussians, and the maximum is set to 4 million Gaussians. 

\paragraph{Evaluation Metrics.} For all methods, we quantitatively assess the diversity and quality of our results by computing the Fréchet Inception Distance (FID) and the recently established DINOv2~\cite{oquab2023dinov2} based Fréchet Distance $\textrm{FD}_{\textrm{DINOv2}}$~\cite{stein2023exposing}. For temporal quality, we use the Fréchet Video Distance (FVD), with a subset of the respective training dataset~\cite{sun2020wod,nuscenes} as reference distribution.
Following \cite{gao2024magicdrive3d}, we evaluate FID and $\textrm{FD}_{\textrm{DINOv2}}$ score on generated results from views seen during the Gaussian Splatting optimization (FID \textit{seen}) as well as from novel views sampled at randomly selected distances from the training ones (FID \textit{novel}).
Additionally, we evaluate prompt adherence using the CLIP score~\cite{Radford2021LearningTV} with the implementation from~\cite{taited2023CLIPScore} on 10 common weather, time-of-day, and localization prompts with 3 samples each.

\paragraph{Quantitative Results.}
We evaluate all methods on a set of 40 generated scenes and across ten different scene attributes (time of day, season, weather).
Exact prompts are provided in the Supplementary material.
As reported in Tab.~\ref{tab:quantitative_results}, the quality of rendered images from our model is on par with state-of-the-art 2D generative methods and is capable of generating scenes closely matching in style and content with the source distribution
However, \emph{competing methods cannot generate a 3D-consistent scene}, resulting in novel views with inferior quality -- as seen in the resulting high FID and FVD score for novel views all other methods.
We also confirm prompt adherence at par with video models, validating that 3D grounding enables fine-grained text control comparable to generalist models~\cite{yu2024wonderjourney}.

\footnotetext[1]{No code publicly available (or from authors). Reimplemented with MagicDriveDiT~\shortcite{gao2024magicdrivedit} and 2DGS~\cite{Huang20242DGS}.}

\footnotetext[2]{We compare scores on the Waymo~\cite{sun2020wod} distributions and released video generations. NuScenes~\cite{nuscenes} results are reported first for completeness of the evaluation.}

\paragraph{Qualitative Results.}
We also find significant qualitative differences between baselines and our method and baselines in Fig.~\ref{fig:qualitative}.
Baseline image-to-video models produce scene renderings of variable quality.
In fact, as views deviate further from the input image, the rendering quality and 3D consistency deteriorate, yielding a Gaussian representation which is inconsistent beyond the original trajectory. 
This difference is more pronounced when departing from the generated video trajectory - even 2.5D methods fail to produce consistent novel views for unlimited viewpoints.
In contrast, our method generates plausible rendering throughout the entire scene, from any realistic viewpoint - without any loss of appearance or geometric quality.

\subsection{Composability with Dynamic Actors}

\begin{figure}[t!]
    \centering
    \includegraphics[width=0.49\linewidth]{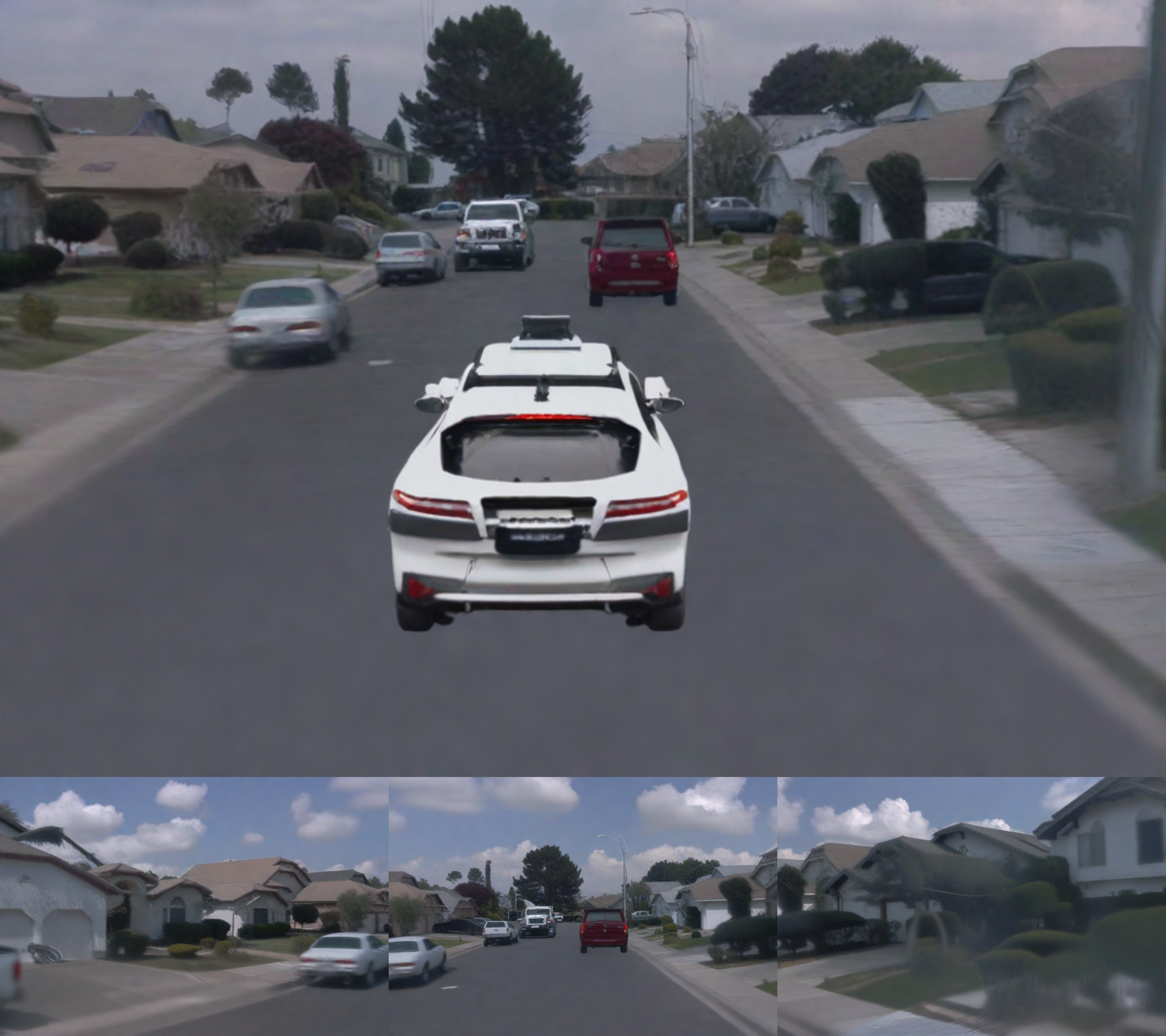}
    \includegraphics[width=0.49\linewidth]{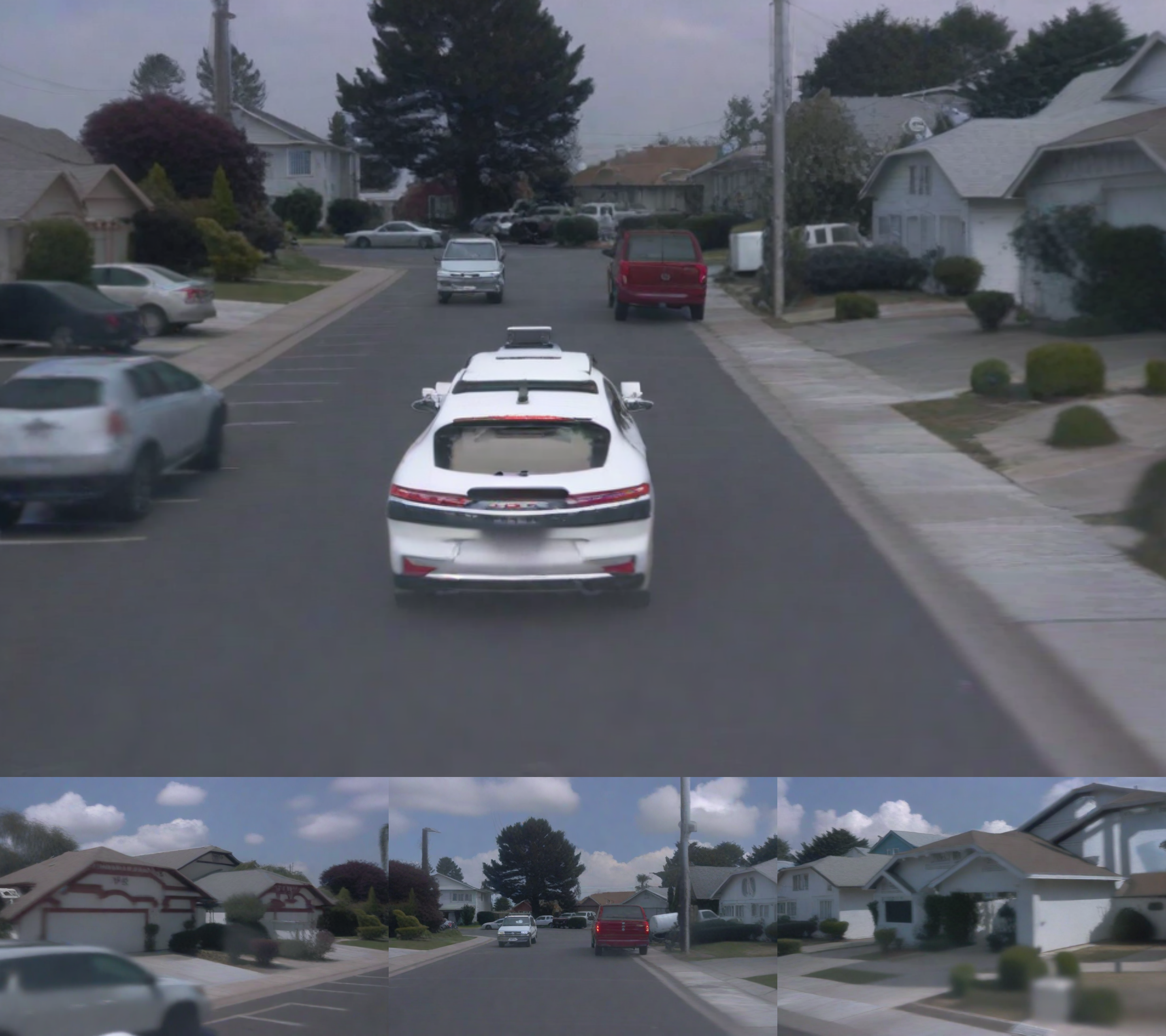}
    \caption{\label{fig:composability}
    \textbf{Composability with Dynamic Actors.} We simulate driving trajectories in a residential street scenes for a Waymo-representative ~\cite{sun2020wod} sensor stack.
    From bottome to top, we show a third-person view of the ego capture vehicle followed by a set of rendered front cameras.
    Both columns correspond to a different timestamp of the same traffic configurations + \emph{fully generated} scene \& asset models.
    The second vehicle is hidden behind the corner in the first frame.
    }
\end{figure}

Being able to place dynamic actors in the generated scene is crucial for downstream applications, including real-time closed-loop simulation.
Our representation directly allows for plug-and-play usage with other components of the simulation stack. We illustrate the integration with actor assets, traffic generation, and sensor stack rendering in Fig.~\ref{fig:composability}.
Using either generated 3D objects, reconstructed objects, or even synthetic objects, the environment map can be used to relight added assets.
In addition, map conditioning directly supports the use of standard traffic generation~\cite{kazemkhani2025gpudrive,dosovitskiy2017carla,li2021metadrive,feng2023trafficgen} and planning modules, providing realistic asset placement and integration in driving scenes, as seen with cars from image-to-3D pipelines~\cite{trellis} placed in our generated scenes.
At each timestamp, we sample object poses from the generated trajectories to place agents within the scene.
These agents are then relit according to the environment map, and the scene is rendered through the ego vehicle’s sensor stack.

\section{Conclusion}\label{sec:discussion}

We introduce, to our knowledge, the first distillation approach to directly generate large-scale explicit 3D driving scenes. To accomplish this, we propose Geometry-Grounded Distillation Sampling (GGDS), which combines controlled proxy mesh generation with a conditional diffusion prior, producing novel and view-consistent Gaussian splatting scenes. Our approach generates completely unseen driving environments controlled by scene descriptions or traffic map layouts. By the design of our method, every scene is generated causally and 3D-consistent, and allows for real-time rendering of physically-grounded videos along novel trajectories. The approach compares favorably against the most successful existing methods -- primarily video diffusion approaches -- that struggle with view-consistent rendering and causality, and are fundamentally limited to individual trajectories. As a geometry-grounded approach, we hope to integrate the method with driving simulators and extend the domain beyond autonomous driving -- ultimately building towards the goal of fully data-driven simulators.

\section*{Acknowledgments} Felix Heide was supported by an NSF CAREER Award (2047359), a Packard Foundation Fellowship, a Sloan Research Fellowship, a Sony Young Faculty Award, a Project X Innovation Award, a Amazon Science Research Award, and a Bosch Research Award. Andrea Ramazzina was supported by the German Ministry BMWE within the project “NXT GEN AI METHODS”.

{
    \small
    \bibliography{main}
}

\end{document}